%% file: main.tex
\documentclass[sigconf]{acmart}
\renewcommand\footnotetextcopyrightpermission[1]{}
\AtBeginDocument{%
  }

\setcopyright{acmlicensed}
\copyrightyear{2018}
\acmYear{2018}
\acmDOI{XXXXXXX.XXXXXXX}
\acmConference[Conference acronym 'XX]{Make sure to enter the correct
  conference title from your rights confirmation email}{June 03--05,
  2018}{Woodstock, NY}
\acmISBN{978-1-4503-XXXX-X/2018/06}

\usepackage[ruled,vlined,noend]{algorithm2e}
\usepackage{enumitem}
\usepackage{multirow}
\usepackage{booktabs}
\usepackage{float}
\usepackage{makecell}
\usepackage{adjustbox}
\usepackage{xspace}

\newcommand{\method}{DevPiolt\xspace}
\newcommand{\stitle}[1]{\vspace*{0.4em}\noindent{\bf #1\/}}
\usepackage{pifont}
\newcommand{\squishlist}{
	\begin{list}{$\bullet$}
		{ \setlength{\itemsep}{1pt}
			\setlength{\parsep}{1pt}
			\setlength{\topsep}{2.5pt}
			\setlength{\partopsep}{0.5pt}
			\setlength{\leftmargin}{1em}
			\setlength{\labelwidth}{1em}
			\setlength{\labelsep}{0.6em}
		}
	}
	\newcommand{\squishend}{
	\end{list}
}
\usepackage{graphicx}
\usepackage{fancyhdr}
\begin{document}

\title{DevPiolt: Operation Recommendation for IoT Devices \\at Xiaomi Home}

\author{
Yuxiang Wang$^{\dagger}$ \quad Siwen Wang$^{\S}$ \quad Haowei Han$^{\dagger}$ \quad Ao Wang$^{\dagger}$ \quad  Boya Liu$^{\S}$ \quad Yong Zhao$^{\S}$ 
\\ Chengbo Wu$^{\S}$ \quad  Bin Zhu$^{\S}$ \quad Bin Qin$^{\S}$ \quad Xiaokai Zhou$^{\dagger}$  \quad  Xiao Yan$^{\dagger}$ \quad Jiawei Jiang$^{\dagger}$ \quad Bo Du$^{\dagger}$
\and
$^{\dagger}$ School of Computer Science, Wuhan University \quad $^{\S}$ Xiaomi
\and
$^{\dagger}$ $\{$nai.yxwang,haowei.han,aowang2024,xiaokaizhou,yanxiaosunny,jiawei.jiang,dubo$\}$@whu.edu.cn \\
$^{\S}$ $\{$wangsiwen,liuboya,zhaoyong6,wuchengbo1,zhubin,qinbin$\}$@xiaomi.com}

\renewcommand{\shortauthors}{Trovato et al.}

\begin{abstract}
Operation recommendation for IoT devices refers to generating personalized device operations for users based on their context, such as historical operations, environment information, and device status. This task is crucial for enhancing user satisfaction and corporation profits. Existing recommendation models struggle with complex operation logic, diverse user preferences, and are sensitive to suboptimal suggestions, limiting their applicability to IoT device operations. To address these issues, we propose \method, an LLM-based recommendation model for IoT device operations. Specifically, we first equip the LLM with fundamental domain knowledge of IoT operations via the pre-training and fine-tuning on delicately constructed device operation datasets. Then, we employ direct preference optimization to align the fine-tuned LLM with specific user preferences. Finally, we design a confidence-based exposure control mechanism to avoid negative user experiences from low-quality recommendations. Extensive experiments show that \method significantly outperforms baselines on all datasets, with an average improvement of 69.5\% across all metrics. \method has been practically deployed in Xiaomi Home app for one quarter, providing daily operation recommendations to 255,000 users. Online experiment results indicate a 21.6\% increase in unique visitor device coverage and a 29.1\% increase in page view acceptance rates.
\end{abstract}



\maketitle

\section{Introduction}

Xiaomi Home, a smart living ecosystem under Xiaomi, primarily oversees the company's smart home business. Xiaomi Home has developed an intelligent service domain encompassing home control, health monitoring, and security protection, using IoT technology to connect and coordinate with terminal devices. 
Xiaomi Home's product line encompasses three main categories: smart home, wearable devices, and smart travel devices, which are further divided into over 260 subcategories, including smart air conditioners (ACs), switches, and curtains. In 2024, its global device shipments reached 60 million units, representing a 20.1\% increase from the previous year, with a cumulative total sales volume of 944 million units. 
To facilitate the management of these smart devices, Xiaomi Home has developed a complementary mobile application (i.e., Xiaomi Home app) dedicated to promoting the in-depth management of IoT devices in daily life scenarios. As of June 2025, the monthly active user base of the Xiaomi Home app reached 113 million, with over 20.5 million users owning five or more smart devices.

\begin{figure}[t]
    \centering
    \includegraphics[scale=0.45]{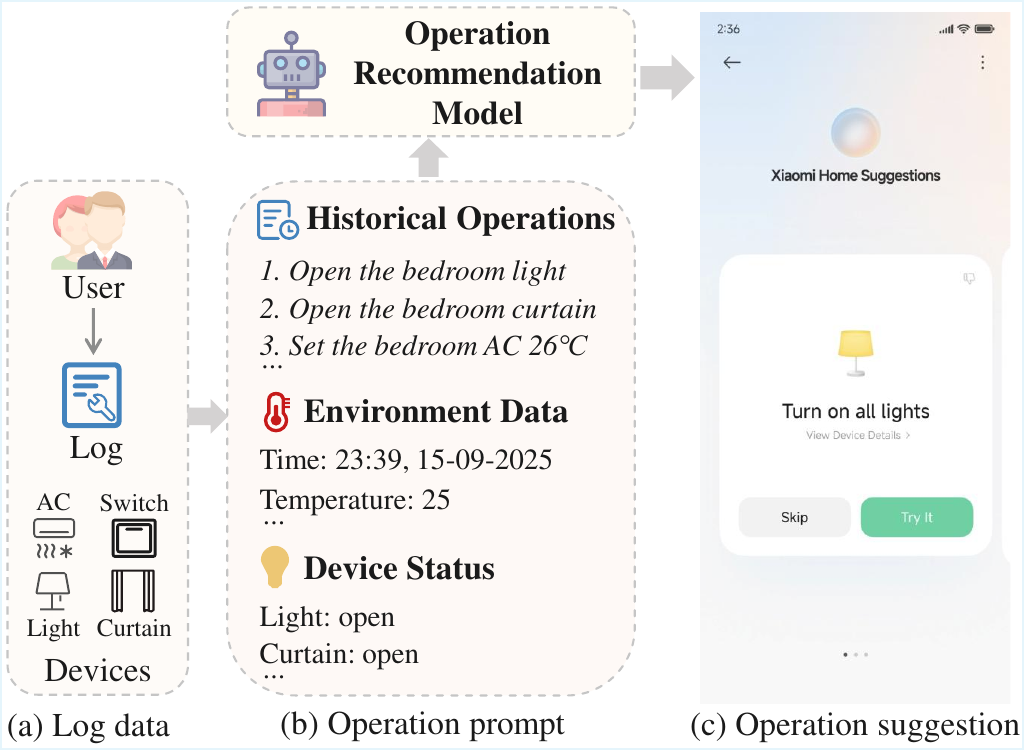}
    \vspace{-1em}
    \caption{The workflow of device operation recommendation in Xiaomi Home app.}
    \vspace{-0.4cm}
    \label{fig:alpha figure}
\end{figure}

Operation recommendation is a core service of Xiaomi Home, designed to provide users with personalized device operation suggestions. This service benefits both users and businesses. 
For users, the increasing variety and functionality of IoT devices make it increasingly inconvenient to achieve precise control based on the current environment. 
Operation recommendation can reduce the frequency of manual operations, thereby enhancing user efficiency and interaction experience. 
For businesses, it allows for the optimization of product design and the increased sales of related devices through user feedback on these recommendations.
As shown in Figure~\ref{fig:alpha figure}, the recommendation engine first collects the user's historical operation sequences (e.g., turning on lights, setting AC temperature), device lists (e.g., lights, ACs), and environmental data (e.g., time, temperature). 
These data are then fed into a recommendation model to predict the user's potential device operation needs.
Finally, it provides operation suggestions to the user in text form. 
For example, if a smart device detects that someone is in the bedroom during nighttime, the operation recommendation predicts that the user may need to ``\textit{close the bedroom curtains}''. 
If the user accepts this suggestion, the Xiaomi Home app will perform the relevant operations via a unified device control interface.

The recommendation model is the core component of operation recommendation and has been extensively studied in both academia and industry~\cite{guo2017deepfm,zhou2018deep,naumov2019deep,zhai2024actions,tang2020progressive}. These models can generally be categorized into two types~\cite{wu2022survey}: Feature-based Recommendation Model (FRM)~\cite{ma2020temporal,tian2021understanding} and LLM-based Recommendation Model (LRM)~\cite{yang2024sequential,yin2024dataset}.
FRMs use carefully designed features as input to construct a mapping relationship between users and items~\cite{zhou2018deep}.
For instance, the Customized Gate Control model~\cite{tang2020progressive} (CGC) dynamically controls the activation of different cross-features through learnable gating parameters, effectively filtering out the most relevant interactions for users. 
LRM methods, such as CALRec~\cite{li2024calrec} and MLLM-MSR~\cite{ye2025harnessing}, transform historical action sequences and item features into natural language, relying on the semantic understanding capabilities of LLMs to uncover the connections between users and items. 
Given the powerful text generation capabilities of LLMs, which can produce suggestions for multiple devices and operations along with corresponding textual descriptions, we develop the operation recommendation based on LLMs.

\begin{table}[t]
\caption{Accuracy comparison results for CGC, DeepSeek and \method on the \textit{Whole} dataset (refer to Section~\ref{sec: Experiment Settings}).}
\vspace{-0.2cm}
\begin{tabular}{l|rrr}
\toprule
\multicolumn{1}{l|}{\multirow{2}{*}{\textbf{Method}}} & \multicolumn{3}{c}{\textbf{Whole}} \\\cline{2-4}
\multicolumn{1}{c|}{}                                 & EM-Acc     & LM-F1     & Rule      \\ 
\midrule
CGC~\cite{tang2020progressive}                                                  & 19.95      & 17.72     & 23.48     \\
DeepSeek~\cite{guo2025deepseek}                                             & 32.97      & 44.68     & 41.45     \\
\method                                               & \textbf{44.33}      & \textbf{57.49}     & \textbf{53.03}    \\
\bottomrule
\end{tabular}
\label{tab: alpha table}
\vspace{-0.3cm}
\end{table}

\stitle{Challenges}. However, when applied to the operation recommendation tasks, Table~\ref{tab: alpha table} shows that current models exhibit low accuracy. 
We identify three key challenges in implementing device operation recommendations using existing recommendation models:

\begin{itemize}[leftmargin=*]
\item[\ding{182}] \stitle{Complex Operation Logic}. Operation recommendations must recall multiple item combinations (e.g., devices and actions). However, existing recommendation models typically handle a single type of item. Furthermore, there is a sequential logic between the operations, such as the need to turn on the power outlet before opening the AC, rather than the reverse.

\item[\ding{183}] \stitle{Diverse User Preferences}. User preferences significantly influence operation recommendations, encompassing both long-term stable preferences (e.g., the habit of opening curtains at a specific time every day) and short-term behavioral patterns (e.g., avoiding frequent toggling of the AC). However, current methods often struggle with diverse user preferences, thereby impeding the effective capture of these specific tendencies.

\item[\ding{184}] \stitle{Sensitive to Inappropriate Suggestions}. The vast pool of candidate operations inevitably includes many suboptimal inappropriate that are illogical and severely degrades the user experience. Therefore, a critical challenge is the prospective identification and filtering of these inappropriate candidates to ensure only high-quality recommendations are presented to the user.
\end{itemize}

To address these challenges, we propose \method, an LLM-based operation recommendation framework, including four core components: \textit{pre-training}, \textit{fine-tuning}, \textit{recommendation refinement} and \textit{exposure control}. 
Specifically, for Challenge \ding{182}, we first pre-train the LLM on a large amount of user historical operations to enable it to learn feasible operation combinations. 
Then, we fine-tune the LLM with fine-grained operation corpus, including historical operations, current environment data and operable devices. The fine-tuning allows LLM to understand logical relationships between operations based on context. Additionally, we pair a textual description with each operation action in the corpus, and prompt the LLM to output both operation action and corresponding descriptions. 
For Challenge \ding{183}, we construct an operation dataset containing positive and negative sample pairs, which are used for direct preference optimization (DPO). This method maximizes the relative likelihood difference between positive and negative samples, enabling the LLM to learn specific user preferences from a mass of historical operations. 
Finally, for Challenge \ding{184}, we introduce an exposure control technique, which determines whether to present operation suggestions to users based on confidence scores, thus avoiding suboptimal recommendations that may lead to a negative user experience.

To train the operation recommendation model, we sample 45,000 operational entries from user logs every three months for pre-training. The fine-tuning corpus contains 15,000 manually annotated samples. The test set comprises 4,882 labeled operations, covering 21 types of IoT devices. We further subdivided the test set based on the device count and device category, resulting in 7 and 17 sub-datasets, respectively.
\method has been practically deployed in the Xiaomi Home app for one quarter, serving 255,000 users daily and generating 326,000 operation suggestions per day.

We evaluate \method and its designs against baselines on various datasets from Xiaomi Home. 
We also introduce three evaluation metrics for the qualitative assessment: exact match accuracy, loose match F1 score and rule score. 
The main results show that \method outperforms the baselines on all datasets. Compared to the best-performing baseline, the average gains in the three metrics are 95.5\%, 58.5\%, and 54.0\%, respectively. 
Ablation studies validate that each proposed technique incrementally improves accuracy when added to a base LLM. Additionally, parameter sensitivity experiments find that model accuracy improves with LLM scale and that setting the context and history sequences to a moderate length yields optimal results. 
Online experiments indicate that \method achieve a 21.6\% increase in unique visitor (UV) device coverage and a 29.1\% increase in page view (PV) acceptance rates.

In this paper, our contributions are summarized as follows:
\squishlist
\item We find that existing methods cannot work well for operation recommendation tasks and identify three major challenges: complex operation logic, diverse user preferences, and sensitive to suboptimal suggestions.
\item We propose \method and design pre-training and fine-tuning to learn operation knowledge on operation datasets \footnote{ We will open-source the operation datasets to enhance further exploration in the community.}, a DPO-based method to reinforce specific user preferences, and an exposure control technique to filter unsatisfied recommendations. 
\item We conduct extensive experiments to evaluate \method. The results show that \method significantly outperforms the baselines, and our designs are effective in improving accuracy.
\squishend

\begin{figure*}[t]
    \centering
    \includegraphics[scale=0.75]{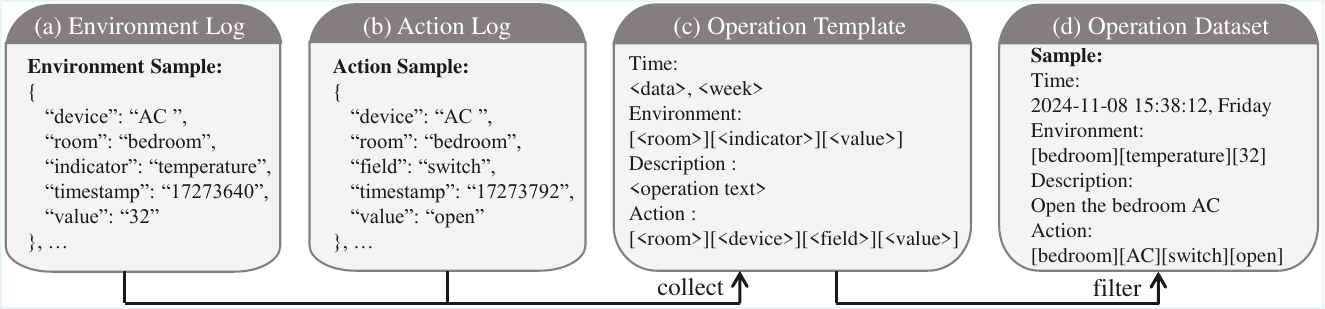}
    \vspace{-0.3cm}
    \caption{The process of constructing operation dataset.}
    \vspace{-1em}
    \label{fig:dataset}
\end{figure*}


\section{Preliminaries}

\stitle{Operation Record Data}. To enable the LLM to learn users' operation preferences, we construct a high-quality dataset of device operation data. As shown in Figure~\ref{fig:dataset} (a) and (b), we aggregate historical operations from the raw environment and action logs of the Xiaomi Home app. Environment logs are the environmental detection data from IoT sensors, such as room temperature and humidity. Action logs are user operation records on smart devices, either through voice or direct control. We record the \textit{device}, \textit{room} and \textit{timestamp} of the operation action, and use \textit{field} and \textit{value} to represent the type and content of operation, respectively.

To translate complex device operation into textual content that LLMs can easily understand, we design an operation template in Figure~\ref{fig:dataset} (c) to filter the log data. 
We denote a device operation dataset $\mathcal{D} = \{\mathcal{H}_1, \mathcal{H}_2,...,\mathcal{H}_N\}$, where $\mathcal{H}$ denotes the user's historical operation sequences and $N$ is the number of users. Furthermore, each sequence  $\mathcal{H}_i=\{\mathcal{O}_1, \mathcal{O}_2, ..., \mathcal{O}_l\}$ includes multiple operations $\mathcal{O}$, where $l$ is the operation count for one user. As shown in Figure~\ref{fig:dataset} (d), each operation consists of the following four components: 
\squishlist
\item \textit{Time} $\mathcal{O}^{time}_i$. We convert the execution timestamp of an action into a date format (i.e., \textit{yy-mm-dd}) and the day of the week.
\item \textit{Environment} $\mathcal{O}^{env}_i$. We construct a triplet $\mathcal{O}^{env}_i
\!=\!\{$[$room$]$,$[$indicator$]$,\\ $[$[value$]$\}$ to describe the current environment.
\item \textit{Description} $\mathcal{O}^{desc}_i$. This is an operation description given by the user through voice or the Xiaomi Home app to the device. 
\item \textit{Action} $\mathcal{O}^{act}_i$. We extract the operation actions executed by users and represent the action content precisely using a quadruplet $\{[room], [device], [field], [value]\}$.
\squishend 

\stitle{The Operation Recommendation Task.} The goal is to generate an action quadruple $\mathcal{O}^{act}_i$ and a description $\mathcal{O}^{desc}_i$ based on the historical operations, current environment, and user's devices. Notably, $\mathcal{O}^{act}_i$ and $\mathcal{O}^{act}_i$ are semantically equivalent in terms of operation meaning.


\input{method}

\begin{table*}[t]
\small
\setlength{\tabcolsep}{4pt}
\renewcommand{\arraystretch}{1.2}
\caption{Accuracy results for operation recommendation. The best results are marked with bold. ``1-5'' denotes the number of devices per user in this sub-dataset ranges from 1 to 5. ``Light'' denotes all the operational devices in this sub-dataset are lights. The same applies to the rest sub-dataset. \textit{Gain} is the average improvement of \method over the best-performing baseline.}
\vspace{-1em}
\centering
\begin{tabular}{cl|*{4}{c}|*{5}{c}|c|c}
\toprule
 &  & \multicolumn{4}{c|}{\textit{Device Count}} & \multicolumn{5}{c|}{\textit{Device Category}} &  &  \\
\cmidrule(lr){3-6} \cmidrule(lr){7-11}
\multirow{-2}{*}{\textbf{Metric}} & \multirow{-2}{*}{\textbf{Method}} & \textbf{1-5} & \textbf{6-10} & \textbf{11-20} & \textbf{20+} & \textbf{Light} & \textbf{Switch} & \textbf{Camera} & \textbf{Warmer} & \textbf{AC} & \multirow{-2}{*}{\textbf{Whole}} & \multirow{-2}{*}{\textbf{Gain} (\%)} \\
\midrule
 & CGC & 43.13 & 20.86 & 13.97 & 7.83 & 19.14 & 18.26 & 33.02 & 27.23 & 12.71 & 19.95 &  \\
 & DeepSeek & 47.53 & 27.50 & 17.26 & 12.44 & 21.67 & 20.90 & 43.71 & 18.32 & 25.16 & 32.97 &  \\
 & GPT-4o & 48.03 & 29.99 & 18.33 & 15.57 & 20.02 & 21.47 & 47.03 & 19.51 & 25.90 & 33.43 &  \\
\multirow{-4}{*}{EM-Acc} & \method & \textbf{68.66} & \textbf{51.69} & \textbf{36.48} & \textbf{27.62} & \textbf{38.56} & \textbf{44.72} & \textbf{77.67} & \textbf{53.00} & \textbf{35.90} & \textbf{44.33} & \multirow{-4}{*}{95.53} \\
\midrule
 & CGC & 49.48 & 22.70 & 16.10 & 7.29 & 13.90 & 15.05 & 33.85 & 32.13 & 17.39 & 17.72 &  \\
 & DeepSeek & 56.38 & 38.19 & 38.10 & 33.23 & 41.66 & 37.46 & 45.27 & 26.42 & 27.37 & 44.68 &  \\
 & GPT-4o & 56.66 & 38.69 & 39.04 & 33.38 & 40.80 & 36.87 & 46.34 & 27.32 & 28.53 & 45.19 &  \\
\multirow{-4}{*}{LM-F1} & \method & \textbf{74.31} & \textbf{61.91} & \textbf{54.81} & \textbf{54.34} & \textbf{55.76} & \textbf{63.78} & \textbf{76.61} & \textbf{58.85} & \textbf{44.61} & \textbf{57.49} & \multirow{-4}{*}{58.49} \\
\midrule
 & CGC & 48.26 & 24.09 & 17.37 & 10.18 & 23.37 & 21.29 & 33.81 & 30.64 & 16.56 & 23.48 &  \\
 & DeepSeek & 55.20 & 40.64 & 33.37 & 23.85 & 35.67 & 35.29 & 54.00 & 31.31 & 35.14 & 41.45 &  \\
 & GPT-4o & 55.61 & 39.14 & 34.26 & 24.04 & 34.96 & 33.94 & 55.19 & 33.26 & 37.29 & 42.32 &  \\
\multirow{-4}{*}{Rule} & \method & \textbf{74.44} & \textbf{59.32} & \textbf{46.23} & \textbf{38.38} & \textbf{55.76} & \textbf{63.78} & \textbf{79.72} & \textbf{60.3} & \textbf{45.18} & \textbf{53.03} & \multirow{-4}{*}{54.04} \\
\bottomrule
\end{tabular}
\label{tab: main result}
\end{table*}

\section{Experimental Evaluations}
\subsection{Experiment Settings}
\label{sec: Experiment Settings}
\stitle{Dataset}. We extract environmental features, device lists, and historical operations from the Xiaomi Home app to construct a dataset for device operation recommendations. 
We sample 45,000 operation instances monthly, with this sampling process conducted continuously over three months for pre-training.
Additionally, we manually annotate the original user logs to generate a high-quality test set consisting of 4,882 instances. Notably, the training and test sets are non-overlapping in time to prevent knowledge leakage into the model training phase. To enhance the diversity of the dataset, we divided it into multiple sub-datasets based on device numbers and categories. A total of 9 datasets are used for evaluation, comprising the 4 sub-datasets with the highest instance counts in each of two above categories, along with the \textit{whole} test set. Detailed dataset statistics are provided in the Appendix~\ref{sec: dataset}.

\stitle{Baselines}. We compare \method with three baselines: 
1) \textbf{CGC Model}. We adapt the customized gate control (CGC) model~\cite{tang2020progressive} to the operation recommendation by making detailed adjustments.
Specifically, we use MLP layers to encode user and device features, and leverage a Transformer to capture operation dependencies in the historical sequence. 
The encoded features are then concatenated and fed together with the target operation into the CGC.
The CGC dynamically controls the activation of different cross-features through learnable gating parameters, effectively filtering out the most relevant target operations with users.
2) \textbf{DeepSeek}. We concatenate the user's historical operation sequence, current environment, and list of operable devices, and then prompt the DeepSeek-V3~\cite{liu2024deepseek} model to directly output the operation recommendation. 
3) \textbf{GPT-4o}~\cite{liu2024deepseek}. Similar to the DeepSeek baseline, we use the GPT-4o model to provide operation suggestions.
Note that current FRMs and LRMs are not designed for device operation recommendation, and thus cannot serve as appropriate baselines.

\stitle{Implementation}. We use Qwen2.5-14B~\cite{yang2025qwen3} as the default model for training and inference. During the pre-training phase, we concatenate multiple historical operation sequences to fill the context window, setting the maximum length to 2048. The model is pre-trained for 1 epoch with a learning rate of 3e-5. In the fine-tuning phase, we set the LoRA rank to 16, LoRA alpha to 32, and LoRA dropout to 0.05. The LoRA~\cite{hu2022lora} update matrices are applied to all projection layers. The model is fine-tuned for 2 epochs with a learning rate of 4e-4. Experiments are conducted on a server equipped with 8 NVIDIA H20 GPUs, each with 96GB of memory.


\stitle{Operation Evaluation Criteria}. 
A reasonable evaluation criterion is essential for the qualitative analysis of operation recommendations. We  denote users' actual actions as ground truth labels $\mathcal{Y}$. The predicted results are represented as $y$. Next, we provide a detailed introduction to our three proposed metrics:

\begin{itemize}[topsep=5pt,leftmargin=*]
\item \textbf{Exact Match Accuracy}. 
An exact matching indicates that the four items of the action quadruplets $\mathcal{O}^{act}$ in the prediction and the actual operations are identical. 

\item \textbf{Loose Match F1 Score}. 
An action may be deemed correct even if it does not exactly match the label, as users can employ various expressions to achieve the same operation intent. A loose match is defined as either the difference between the predicted and actual action value is below 20\%, or the predicted and actual actions being functionally equivalent.
For example, the predicted action ``\textit{Set bedroom AC to comfort mode}'' is considered equivalent to the actual one ``\textit{Turn on the bedroom AC}''. We provide detailed explanations in Appendix~\ref{sec: loose match}. We calculate the loose match F1 score based on this loose matching criterion:
\abovedisplayskip=5pt \belowdisplayskip=5pt
\begin{equation}
    \mathrm{LM\text{-}F1}=2\cdot \frac{lm\_precision\cdot lm\_recall }{lm\_precision + lm\_recall }.
    \nonumber
\end{equation}

\item \textbf{Rule Score}. 
The rule score assesses the accuracy and comprehensiveness of the model's recommendation results in multi-device, multi-operation scenarios. 
We consider the single action quadruplet setting, which includes four metrics: \textit{identical, intersecting, containing}, and \textit{opposite}:
\begin{equation}
    \varsigma =\begin{cases}
 1, & \mathcal{Y}=y\\
 1/|E|,  & E=\mathcal{Y}\cap y\\
 0.9, & \mathcal{Y}\subset  y\\
 -0.1, & y=-\mathcal{Y}
\end{cases},
\nonumber
\end{equation}
where $-\mathcal{Y}$ indicates the opposite operation. The total rule score is the sum of the scores for each action quadruple instance, divided by the number of predicted quadruples $N_{quad}$: $\mathrm{Rule}=\left (  {\textstyle \sum_{i=1}^{N_{quad}}} \varsigma_i \right )   /{N_{quad}}$.
This metric is reasonable as it considers both the accuracy of individual operations and the overall accuracy across multiple operations.
\end{itemize}

\subsection{Main Results}

We compare \method with CGC model, DeepSeek and GPT-4o to evaluate the accuracy of recommendation results. Table~\ref{tab: main result} reports the experimental results on 9 sub-datasets and the \textit{whole} dataset. We have following three observations: 1) \method outperforms the baselines on all three accuracy metrics across all datasets, demonstrating a maximum average of 95.53\%, 58.49\%, and 54.04\% over the best-performing baseline. \method surpasses CGC model by leveraging the context understanding capability of the LLM to generate logically coherent operation combinations. Additionally, \method outperforms GPT-4o because the LLM learns specific operation patterns and user preferences through pre-training, fine-tuning, and DPO. 2) As the number of devices increases, the accuracy of all methods decreases, due to the increased complexity in making accurate predictions. 3) Our method achieves the highest accuracy on the Camera dataset but relatively lower accuracy on the Light dataset. This is because a camera has only two possible operations (i.e., on and off), while lights may have more complex operations, including lighting modes and brightness control.
    
\begin{table}[t]
\caption{Ablation study of fine-tuning, pre-training and recommendation refinement for \method. \textit{Base} is the naive Qwen2.5-14B model.}
\begin{tabular}{l|ccc}
\toprule
\multirow{2}{*}{\textbf{Method}} & \multicolumn{3}{c}{\textbf{Whole}} \\ \cline{2-4}
                                 & EM-Acc     & LM-F1     & Rule      \\
                                \midrule
Base                             & 32.04      & 44.05     & 41.92     \\
+ Fine-tuning                    & 42.36      & 56.12     & 51.23     \\
+ Pre-training                   & 43.57      & 56.99     & 52.74     \\
+ Recommendation refinement                            & 44.33      & 57.49     & 53.03    \\
\bottomrule
\end{tabular}
\label{tab: ablation study}

\end{table}

\subsection{Micro Experiments}
\label{sec: micro experiments}

\stitle{Ablation Study}. To study the impact of each component in \method, we gradually activate \textit{fine-tuning}, \textit{pre-training}, and \textit{recommendation refinement} on the base LLM (i.e., Qwen2.5-14B) and evaluate them on the whole dataset. As shown in Table~\ref{tab: ablation study}, each module significantly improves the model's accuracy, demonstrating the effectiveness of our techniques. Fine-tuning allows the LLM to learn fine-grained IoT device operation patterns, pre-training injects domain-specific knowledge by leveraging a large amount of historical operation sequences, and recommendation refinement ensures the LLM learns specific user preferences.

\begin{figure}
    \centering
    \includegraphics[width=0.9\linewidth]{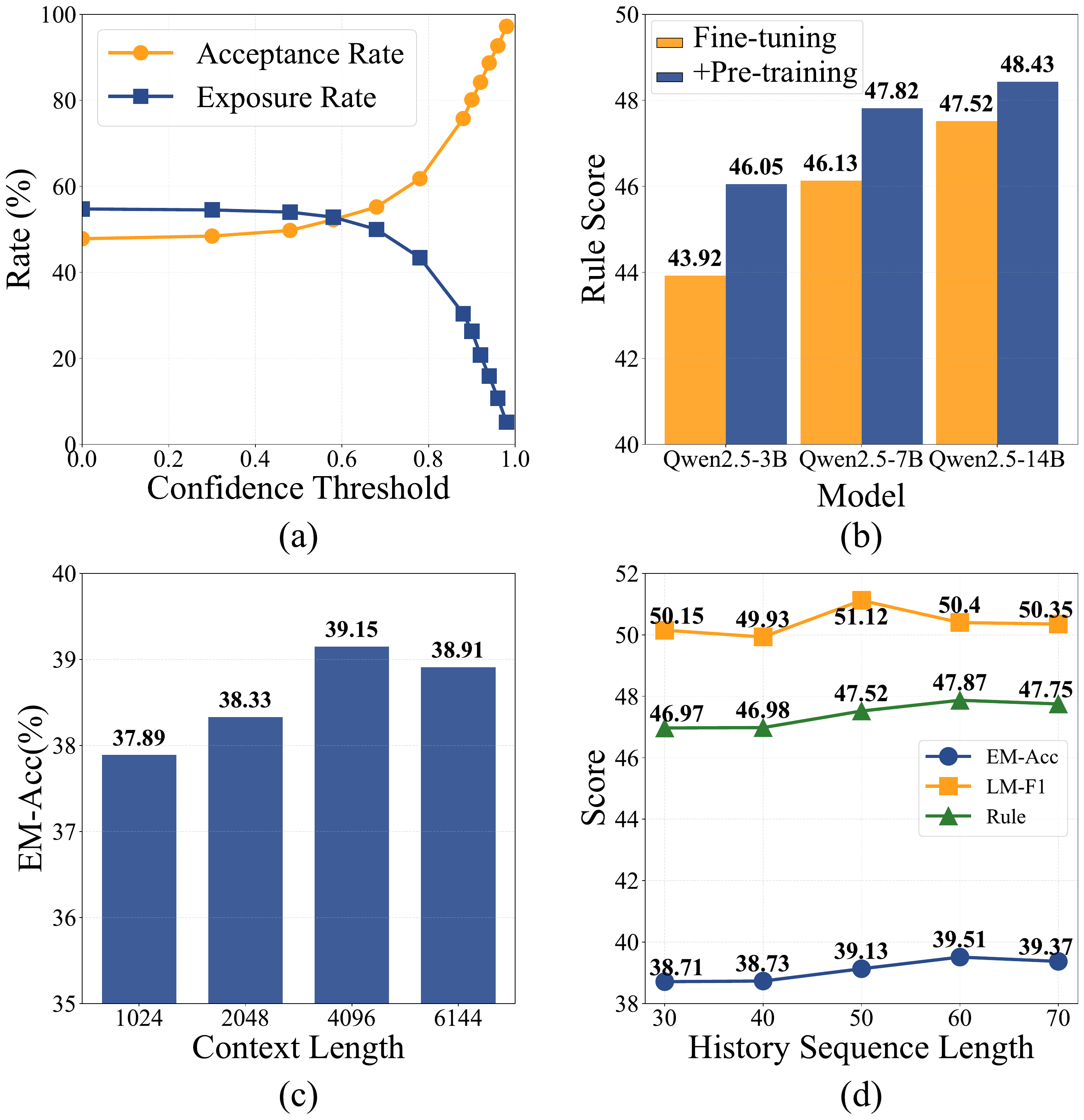}
    \vspace{-1em}
    \caption{The comparison results of model accuracy with different parameters.}
    \label{fig:parameter comparison}
    \vspace{-1mm}
\end{figure}

\stitle{Evaluation of Exposure Control}. To investigate the role of exposure control, we gradually adjust the confidence level from 0 to 1 and observe the changes in exposure rate and acceptance rate. Acceptance rate refers to the proportion of operation suggestions accepted by users out of the total number of operation suggestions. The exposure rate indicates the proportion of suggestions shown to users out of the total number of generated suggestions. As shown in Figure~\ref{fig:parameter comparison} (a), as confidence increases, the exposure rate initially decreases slowly and then more rapidly, while the acceptance rate shows the opposite trend. Higher confidence indicates a more conservative recommendation strategy, thus reducing the likelihood of presenting prediction results to users. However, higher confidence also means the model is more certain in its predictions, leading to a higher acceptance rate. To balance the trade-off between exposure rate and acceptance rate, we set the confidence threshold to 0.7.

\stitle{Action-First vs. Text-First}. We evaluate the accuracy impact of two operation generation strategies on the whole test set: action-first and text-first. The comparison results are provided in Table~\ref{tab: action or text}. We observe that the action-first strategy outperforms the text-first strategy on all three metrics. This is because action quadruples provide a more fine-grained representation of operations compared to textual descriptions. The detailed action quadruple can help model to improve device recall. Conversely, if rough operation descriptions are generated first, the action quadruples may converge to the description due to the LLM's next token generation mechanism, thereby harming accuracy.

\stitle{Impact of Different Model Sizes}. We evaluate the impact of different Qwen2.5 model sizes on accuracy across the whole test set. ``Fine-tuning'' refers to the fine-tuning to the original Qwen model, whereas ``+Pre-training'' indicates fine-tuning on a pre-trained Qwen model. As shown in Figure~\ref{fig:parameter comparison}(b), we observe that accuracy increases with the model size, as larger models exhibit stronger expressive power and a better ability to model long historical operation sequences. Additionally, incorporating a pre-training stage into smaller models can achieve comparable accuracy to that of larger models. This suggests that pre-training learns domain-specific knowledge, and thus improving model accuracy.

\stitle{Impact of Context Length and Historical Sequence Length}. We study the impact of context length during training and the length of historical sequences during testing. The comparison results are presented in Figure~\ref{fig:parameter comparison}(c), (d). We find that very short context lengths limit the operational knowledge the model can learn, while excessively long context lengths cause the model to lose focus. A similar phenomenon is observed with the length of historical sequences. Therefore, a moderate context length and historical sequence length are crucial for optimal model accuracy.

\begin{table}[t]
\caption{The comparison of model accuracy between action-first and text-first strategies. "Action-first" indicates that actions are generated before text during fine-tuning, whereas "Text-first" follows the reverse order.}
\begin{tabular}{l|ccc}
\toprule
\multirow{2}{*}{\textbf{Method}} & \multicolumn{3}{c}{\textbf{Whole}} \\ \cline{2-4}
                        & EM-Acc     & LM-F1      & Rule  \\ 
                        \midrule
Action-first            & 44.33 & 57.49 & 53.03 \\ 
Text-first              & 42.43 & 56.66 & 51.75 \\ 
\bottomrule
\end{tabular}
\label{tab: action or text}
\end{table}

\stitle{Failure Analysis}.
We randomly sample 50 user-disliked operation suggestions over a week to conduct a failure analysis. 
We categorize the failure causes into six types: 
\squishlist
\item  \textit{Incomplete operations}. The generated suggestion fails to cover all user's actual operations. 
\item \textit{Unclear description}. The generated operation descriptions are difficult for users to understand (e.g., a suggestion to "turn off the light" without specifying the room).
\item \textit{Imperceptible environment}. Inadequate environmental data leads to inaccurate operation suggestions (e.g., suggesting to turn off the lights at night when someone is in the room).
\item \textit{Inappropriate timing}. The suggested operations are often ill-timed (e.g., turning on all lights during the daytime).
\item \textit{Lack of historical operations}. The model's recommendations are inaccurate due to the limited historical operations. (Direct user control of devices results in missing operation records.)
\item \textit{Unreasonable recommendations}. The operation suggestions can not align with the user's needs (e.g., turning off the AC when the room temperature is high).
\squishend

\begin{figure}
    \centering
    \includegraphics[width=1\linewidth]{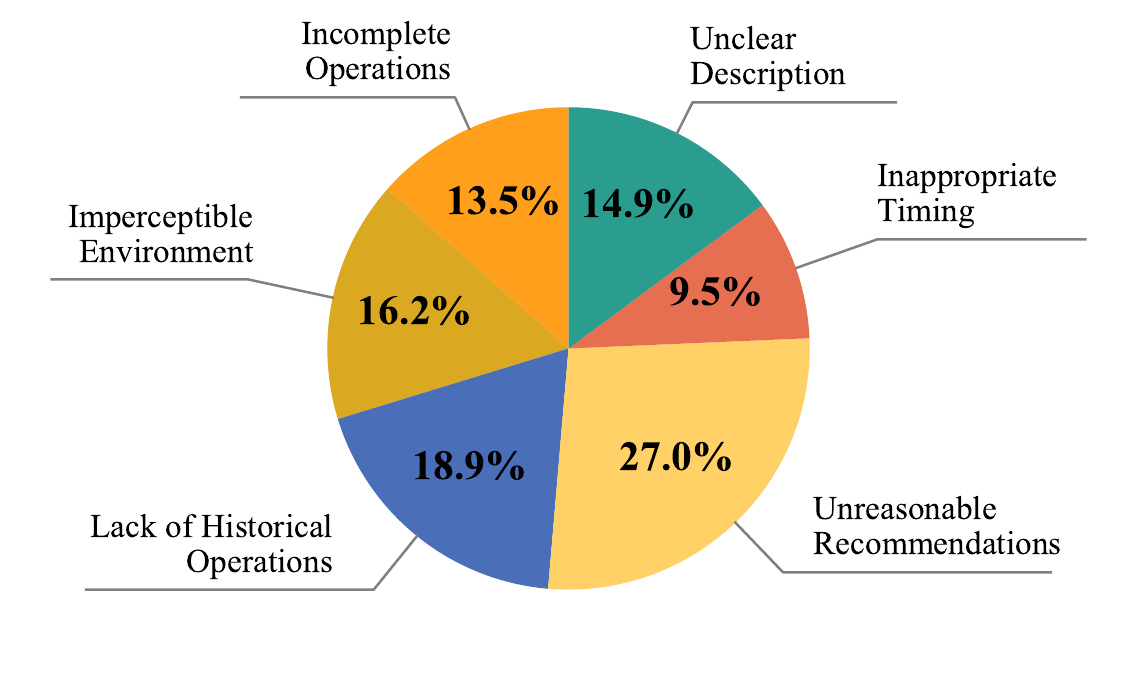}
     \vspace{-4em}
    \caption{Case study for user-disliked operation suggestions.}
    \label{fig:case study}
\end{figure}

Figure~\ref{fig:case study} shows that over 45\% of the failed cases are due to unreasonable recommendations and lack of historical operations. This highlights the importance of enriching historical operational data and using it to better understand user intentions, which is crucial for the accuracy of the recommendation model.
The fewest failures are due to inappropriate timing, suggesting that our recommendation refinement technique can effectively mitigate such issues. 
The remaining three categories have roughly the same number of cases, indicating that the model still faces significant challenges in complex environments with multiple devices and operations.

\section{Related Work}
\stitle{Recommendation Models}.
Existing recommendation models can be categorized into two types: feature-based recommendation models (FRMs)~\cite{wang2017deep,guo2017deepfm,naumov2019deep,zhai2024actions,kang2018self} and LLM-based recommendation models (LRMs)~\cite{zhang2025recommendation,hu2024enhancing,liu2024llm,harte2023leveraging,li2023prompt}. Specifically, for FRMs, DIN~\cite{zhou2018deep} captures users' interest features using an attention mechanism and deep neural networks, and employs MLP to model high-order interactions between features. DirectPred~\cite{tian2021understanding} distinguishes between short-term behavioral preferences and long-term interest features through contrastive learning. HRNN~\cite{ma2020temporal} models the temporal dependencies of users' real-time behavior and contextual features using causal and dilated convolutions in a temporal convolutional network. For LRMs, CALRec~\cite{li2024calrec} dynamically aligns users' behavior sequences with target item features in the latent space through fine-tuning and multi-task learning based on an LLM. Recformer~\cite{li2023text} extracts item text features using a pre-trained language model and models user action sequences with a Transformer, achieving the mapping between user intent and item semantics using only textual modality. LRURec~\cite{yue2024linear} achieves fast training and dynamic interest modeling with a simplified linear recurrent structure and parameter sharing, thus enhancing the sequential prediction capabilities while maintaining computational efficiency. However, none of these methods can be applied to operational recommendations, as they are unable to capture complex operational logic and patterns.

\stitle{Management for IoT Devices.}
IoT device management~\cite{yu2019deep,braten2020autonomous,perumal2015iot,bajpayi2024ai,haque2024scalable} encompasses the configuration and scheduling of IoT devices to ensure their efficient and reliable operation. AutoIOT~\cite{shen2025autoiot} achieves automated AIoT application management by using LLM-driven natural language programming, which generates executable configuration instructions for devices through semantic parsing. qToggle~\cite{srinivasrao2024user} designs a low-power communication protocol and an embedded sensor network to enable interconnectivity and autonomous decision-making among IoT devices. Cristina et al.~\cite{stolojescu2021iot} employs a cost-effective Wi-Fi module as the core controller and integrates physical switches to unify local and remote control in smart homes. However, these methods are limited to basic device control and can not offer fine-grained operation suggestions based on the current environment.

\section{Conclusion}
We introduce \method, an LLM-based operation recommendation for IoT devices. 
Specifically, we pre-train and fine-tune LLMs to understand the basic logic of device operations. We then use direct preference optimization to tailor the suggestions to individual user preferences. Finally, we implement a confidence-based exposure control technique to prevent the presentation of suboptimal suggestions. 
Extensive experiments demonstrate that our approach outperforms all baselines, paving the way for community research in operation recommendations.

\bibliographystyle{ACM-Reference-Format}
\bibliography{sample-base}

\appendix
\raggedbottom
\section{Dataset}
\label{sec: dataset}
We provide the statistical data of the test set in Table~\ref{tab: dataset statistics}. ``DC 1-5'' indicates that the number of user devices in this sub-dataset is between 1 and 5, and ``Light'' indicates that all the objects being operated on in this sub-dataset are lights. The remaining datasets follow a similar pattern. \textit{Whole} represents the entire dataset. The numbers in the table indicate the number of samples included in each sub-dataset.

\begin{table*}[!htb]
\centering
\caption{Test Set Statistics. \#Samples denote the number of samples.}
\begin{tabular}{lllllllllll}
\toprule
\textbf{Dataset} & \textbf{DC 1-5}         & \textbf{DC 6-10}          & \textbf{DC 11-20}         & \textbf{DC 20+}           & \textbf{Light}            & \textbf{Switch}           & \textbf{Camera}         & \textbf{Warmer}         & \textbf{AC}             & \textbf{Whole}            \\
\midrule
\#Samples        & \multicolumn{1}{r}{970} & \multicolumn{1}{r}{1,037} & \multicolumn{1}{r}{1,650} & \multicolumn{1}{r}{1,225} & \multicolumn{1}{r}{1,211} & \multicolumn{1}{r}{1,599} & \multicolumn{1}{r}{318} & \multicolumn{1}{r}{200} & \multicolumn{1}{r}{390} & \multicolumn{1}{r}{4,882}\\
\bottomrule
\end{tabular}
\label{tab: dataset statistics}
\end{table*}

\section{Details for Loose Match Criterion}
\label{sec: loose match}
The loose matching criterion comprises two necessary conditions. First, the difference between the predicted and actual continuous parameter values must be within 20\%. Specifically, if the operation value is numerical (e.g., temperature setting or device power level), we calculate the relative error using the formula:
\begin{equation}
    \epsilon  = \frac{|y_p - y_a|}{y_a},
    \nonumber
\end{equation}
 where $y_p$ is the predicted value and $y_a$ is the actual value. The condition for loose matching is satisfied when $\epsilon  \le  0.2$. Second, the operational intent must be functionally equivalent. For instance, if the actual operation is ``Turn on bedroom AC'', a model prediction of ``Set bedroom AC to 26'' is also considered correct. Similarly, ``Activate living room light'' and ``Adjust living room ceiling light to 50\% brightness'' are both deemed to achieve equivalent lighting effects.

\end{document}

%% file: method.tex
\begin{figure}[t]
    \centering
    \includegraphics[scale=0.5]{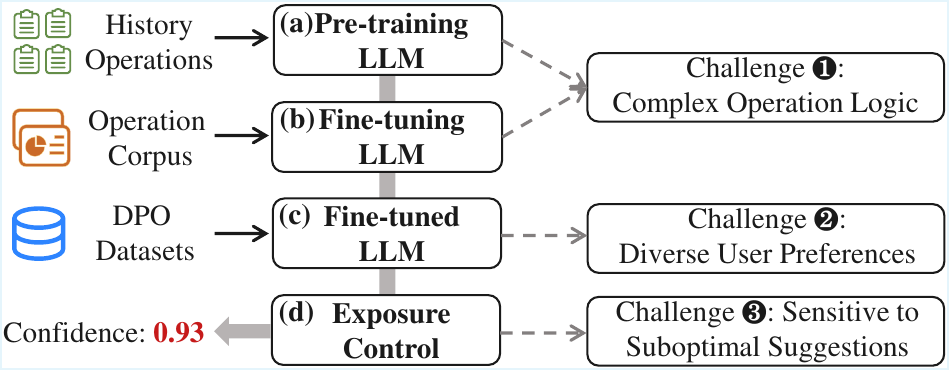}
    \vspace{-1em}
    \caption{The overview of \method.}
    \vspace{-1em}
    \label{fig:overview}
\end{figure}

\section{The \method Framework}
\stitle{Overview}.
To address the four challenges faced in operation recommendations, we propose \method, which comprises four modules: \textit{pre-training, fine-tuning, recommendation refinement}, and \textit{exposure control}. As shown in Figure~\ref{fig:overview}, for challenge 1, we first concatenate the user's historical operation sequences to pre-train an LLM, and then fine-tune the pre-trained LLM using a fine-grained operation corpus, enabling the model to understand the logic of the operations and generate reasonable suggestions. For challenge 2, we further utilize direct preference optimization to refine the fine-tuned LLM. This technique allows the model to learn the specific user preferences. Finally, for challenge 3, we implement exposure control by calculating a confidence score, which determines whether to display the recommendation results to the user.

\begin{figure}
    \centering
    \includegraphics[width=\linewidth]{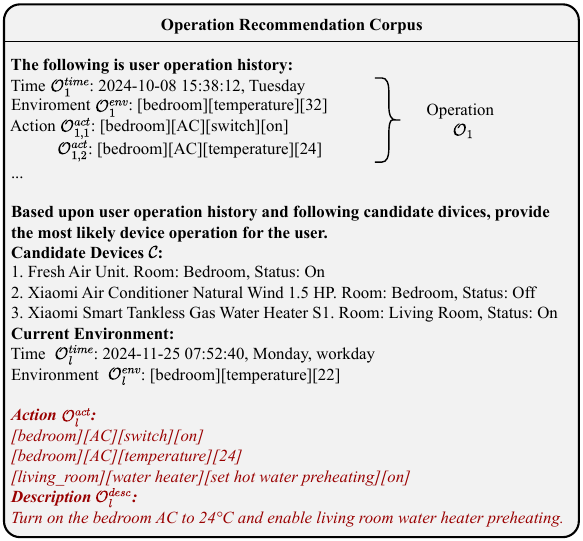}
    \vspace{-2em}
    \caption{An operation dataset sample. The black text represents the operation recommendation prompt $\mathcal{P}$, and the italicized part indicates the target used for loss calculation. The pre-training only includes operation history, while fine-tuning additionally includes environment and device data.}
    \vspace{-0.5cm}
    \label{fig:finetune}
\end{figure}

\subsection{Pre-training and Fine-tuning}
    As highlighted in the challenge 1, while LLMs can leverage their vast world knowledge to reason about intricate relationships within operations, their performance is suboptimal due to a lack of domain knowledge in IoT device operations. Furthermore, directly employing a naive LLM for the recommendation task does not guarantee that its output will conform to a legal operation structure.

    \stitle{Pre-training.}
        To ensure the LLM is able to understand the device operation, we pre-train the base LLM on user  historical sequence $\mathcal{H} = \{\mathcal{O}_1, \mathcal{O}_2, ..., \mathcal{O}_l\}$. As shown in the ``user operation history'' part of Figure~\ref{fig:finetune}, each operation instance $\mathcal{O}_i$ is serialized into a single textual sequence that includes the time $\mathcal{O}^{time}_i$ and environment $\mathcal{O}_{i}^{env}$, followed by the corresponding $a_i$ action(s) $\mathcal{O}_{i}^{act} = \{\mathcal{O}_{i,1}^{act}, ..., \mathcal{O}_{i,a_i}^{act}\}$.
        For instance, in $\mathcal{O}_1$, the user performs logical actions: turning on the AC and setting its temperature to 24 degrees when the bedroom temperature reaches 32 degrees.
        Our training approach reframes the task as next-action prediction, which is a natural fit for LLMs. We concatenate the user's historical actions $\mathcal{O}_{<i}$, along with current time $\mathcal{O}_i^{time}$ and environmental information $\mathcal{O}_i^{env}$, as conditional inputs for the model to predict the user's next action.
        This is formally achieved by minimizing the negative log-likelihood loss, a standard objective for training LLMs:
        \abovedisplayskip=4pt  
        \belowdisplayskip=4pt  
         \begin{equation}
            \mathcal{L}_{pt} = - \sum_{\mathcal{O}_i \in \mathcal{H}} 
            \log \text{P}(\mathcal{O}_{i}^{act} |\mathcal{O}_{<i}, \mathcal{O}_i^{time}, \mathcal{O}_i^{env})
        \end{equation}
        By learning to predict actions from context, the model internalizes the logic of device operations. For example, it learns device-specific functionalities (e.g., a light can only be on or off, not set to 25 degrees) and valid value spaces (e.g., an AC's temperature must be a number within a reasonable range).
        Finally, to ensure the model retains its general knowledge and avoids catastrophic forgetting, we create a balanced training dataset by mixing our operation pre-training corpus with a general-domain corpus, such as ShareGPT~\cite{chen2024sharegpt4v} and WuDao~\cite{yuan2021wudaocorpora}, at a 1:1 ratio.

    After the pre-training, the model is equipped to understand the domain of IoT operations. However, two key gaps remain. 1) The model \textit{struggles to apply its operational knowledge to recommendations}, such as by failing to suggest actions for the user's available devices. 2) It is \textit{unable to generate user-friendly natural language descriptions} (e.g., ``Turn on the AC'') to present to the user instead of the raw structured operations (e.g., [AC][switch][on]).

    \stitle{Fine-tuning.}
        To bridge the two gaps, we fine-tune the LLM on a specialized training corpus, which contains 15,000 manually annotated operation instances. Each training instances consists of an operation recommendation prompt, the ground-truth action(s) and a corresponding textual description. As shown in Figure~\ref{fig:finetune}, we construct the operation recommendation prompt $\mathcal{P}$ by the user's operation history (excluding the final operation) $\mathcal{O}_{<l}$, along with the current time $\mathcal{O}_l^{time}$ and environment information $\mathcal{O}_l^{env}$ and a list of candidate devices $\mathcal{C}$, which is included to train the model to recommend valid operations based on the available devices.
        We fine-tune the LLM using LoRA~\cite{hu2022lora}. It introduces low-rank matrices for incremental parameter updates in the pre-trained LLM, thereby training only a small number of learnable parameters. 

        We fine-tune the model using a multi-task objective to predict both the final action(s) $\mathcal{O}_l^{act}$ and the corresponding description $\mathcal{O}_l^{desc}$. Our initial approach is to generate both outputs simultaneously, by minimizing the following joint probability objective:
        \begin{equation}
            \mathcal{L}_{ft}' = -  \log \text{P}(\mathcal{O}_{l}^{act}, \mathcal{O}_l^{desc} |\mathcal{P})
        \end{equation}
        However, we find this method yield suboptimal accuracy, as the model struggle to master both tasks at once. Therefore, the model first predicts the action and then generates the description. This factorizes the action-description objective $\mathcal{L}_{ad}$ as follows:
        \begin{equation}
            \mathcal{L}_{ad} = - (\log \text{P}(\mathcal{O}_{l}^{act} |\mathcal{P}) + \log \text{P}(\mathcal{O}_{l}^{desc} |\mathcal{P}, \mathcal{O}_{l}^{act}))
        \end{equation}

        \stitle{Insight.}
        We adopt an \textit{action-first} strategy, generating the action before the description, to establish a more logical, concrete-to-abstract learning flow. Predicting the structured and precise action serves as the core task, providing a clear and unambiguous ``anchor'' for subsequent text generation. Once the action is determined, generating the description becomes a simpler and more grounded summarization task. In contrast, a \textit{text-first} strategy, which generates the free-form description first, would require the model to infer a precise action from potentially vague or incomplete text. This process is more challenging and error-prone. Consequently, our approach ensures the model prioritizes mastering the core logic of the recommendation, thereby enhancing overall accuracy and reliability. The comparison results are presented in Section~\ref{sec: micro experiments}.

\begin{figure}
    \centering
    \includegraphics[width=1\linewidth]{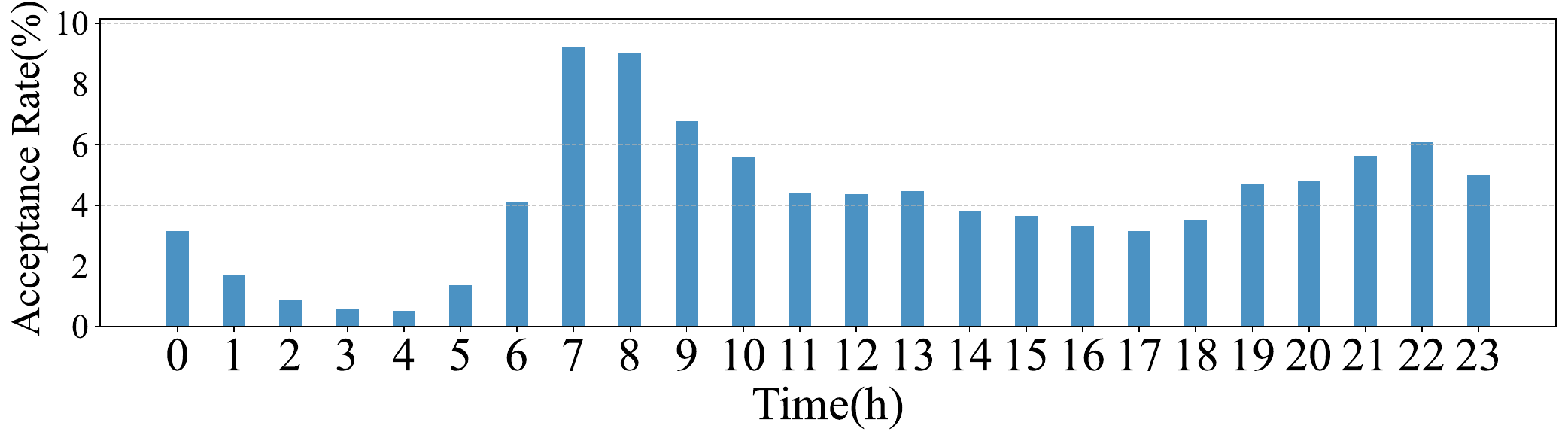}
    \vspace{-2em}
    \caption{The acceptance rate of curtain operation recommendations by users at different times of the day.}
    \label{fig: dpo curtain}
\end{figure}

\subsection{Recommendation Refinement}
    Although the model captures the general distribution of IoT operations by training on extensive user history, its performance can be hindered by special scenarios that fall outside this distribution. We observe that such scenarios are typically caused by specific user preferences, which can be broadly classified into two categories:

    \squishlist
        \item \stitle{Time-sensitive Operations}: Preferences for certain operations can be strongly time-dependent. As shown in Figure~\ref{fig: dpo curtain}, the smart curtain exemplifies this pattern: recommendations for its operation are often accepted in the morning and at night but rejected during the day. Our analysis shows that such recommendations often fail when the recommended operations is inconsistent with a user's historical behavior within a specific time window.

        \item \stitle{Conflicting Operations}: Users are unlikely to perform contradictory actions in quick succession, leading to the rejection of conflicting recommendations. For example, suggesting to turn on the air conditioner moments after the user has manually switched it off would be considered intrusive and is typically rejected.

    \squishend

    To enable our recommendation model to learn user preferences under these special scenarios, we refine it using Direct Preference Optimization (DPO)~\cite{rafailov2023direct}. The core principle of DPO is to directly optimize the model on preference data. It uses pairs of preferred (positive) and dispreferred (negative) outputs to increase the likelihood of the positive samples while decreasing that of the negative ones. Compared to traditional reinforcement learning methods, the DPO sidesteps the need for an explicit reward model, resulting in a simpler and more stable training process.

\begin{figure}
    \centering
    \includegraphics[width=\linewidth]{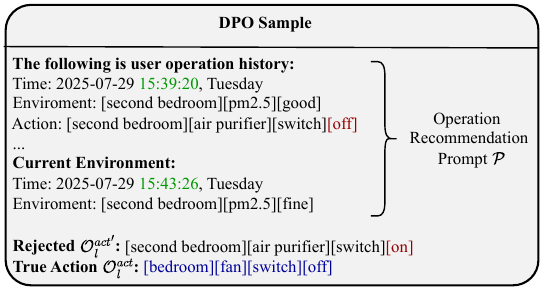}
    \vspace{-2em}
    \caption{An example of a DPO sample. Note that the text in red represents conflicting operations, and text in green indicates similar timestamps.}
    \label{fig:dpo}
\end{figure}

    \stitle{DPO Sample Construction.}
        For DPO training, we construct a dataset from a two-week sample of user data, comprising approximately 9,000 interactions. From this data, we build user preference pairs, each containing a positive sample $S^{pos}$ and a negative sample $S^{neg}$. As shown in Figure~\ref{fig:dpo}, the positive sample $S^{pos}$ consists of the operation recommendation prompt $\mathcal{P}=\{\mathcal{O}_{<l}, \mathcal{O}_l^{time}, \mathcal{O}_l^{env}\}$ and the user's actual executed action $\mathcal{O}_l^{act}$. The negative sample $S^{neg}$ leverages the same prompt $\mathcal{P}$ but is paired with a dispreferred action $\mathcal{O}_l^{act'}$. These pairs formally defined as:
        \begin{equation}
            S^{pos} = \{\mathcal{P}, \mathcal{O}_l^{act}\}, \quad S^{neg} = \{\mathcal{P}, \mathcal{O}_l^{act'}\}
        \end{equation}
        
        Here, the dispreferred action is identified based on the two scenarios described above:

        \squishlist
            \item \stitle{Time-sensitive Rejection.} An action is labeled as dispreferred ($\mathcal{O}_l^{act'}$) if the user has no history of performing it within a ±1 hour window of the recommendation time, thus flagging it as temporally inappropriate.

            \item \stitle{Conflicting Rejection.} An action is considered dispreferred ($\mathcal{O}_l^{act'}$) if it is the inverse of an action the user performed on the same device within the last 10 minutes. To avoid misclassifying common high-frequency interactions (e.g., toggling a light), this time window is shortened to two minutes for such operations.
        \squishend

   \stitle{DPO Training}. We train the model by optimizing the log-probabilities of these sample pairs. The DPO objective is formalized as:
    \begin{equation}
        \mathcal{L}_{DPO} \!=\! - \log \sigma \left( \beta \log \frac{\text{P}_{\theta}(\mathcal{O}_l^{act} | \mathcal{P})}{\text{P}_{ref}(\mathcal{O}_l^{act} | \mathcal{P})} \!-\!\beta \log \frac{\text{P}_{\theta}(\mathcal{O}_l^{act'} | \mathcal{P})}{\text{P}_{ref}(\mathcal{O}_l^{act'} | \mathcal{P})} \right)
    \end{equation}
    where $\text{P}_{\theta}$ denotes the probability computed by the current model, $\text{P}_{ref}$  is the probability from the reference model (i.e., the fine-tuning stage), and $\beta$ is a hyperparameter that controls the strength of the optimization. Compared to the reference model, this objective function encourages models to assign higher probabilities to preferred actions and lower probabilities to dispreferred actions.

\subsection{Confidence-based Exposure Control}
    In operation recommendation tasks, users are sensitive to suboptimal suggestions, which severely degrades the user experience. This prompts us to ask a fundamental question: \textit{How can we assess recommendation quality and control exposure accordingly?} To answer this question, we analyze rejected recommendations and find that users are significantly more likely to reject operations that the model predicts with low confidence. To improve the operation suggestion quality, we adopt a confidence-based exposure control mechanism. Specifically, we expose a suggestion only when the model's confidence in the prediction exceeds a certain threshold. This confidence is calculated as a weighted average of the model's output probabilities for the key attributes of the suggestion.
    
    \stitle{Confidence Score}. We compute the confidence score $Conf(\mathcal{O}_i)$ for each recommended operation. As an operation consists of $a_i$ actions, the final score is an average across them. For each action $\mathcal{O}_{i,j}^{act}$, we focus on the model's predicted probabilities for its key attributes, which are drawn from a vocabulary $\mathcal{V} = \{\text{device}, \text{field}, \text{value}\}$ (e.g., [AC][temperature][24] in the labeled action within Figure~\ref{fig:finetune}). To reflect their relative importance, the probability for each attribute type $k$, denoted as $\text{P}(\text{attr}_k(\mathcal{O}_{i,j}^{act})|\mathcal{P})$, is scaled by a predefined weight, $\alpha_k$. Then, the total confidence value is calculated by summing up these weighted probabilities of all attributes across all actions and normalizing the value to [0,1] by dividing it by the total number of actions.
    The process can be defined as:
    \begin{equation}
        Conf(\mathcal{O}_i)= \frac{1}{a_i} \sum_{\text{attr}_k \in \mathcal{V}} \sum_{j=1}^{a_i}
        \alpha_{k}\text{P}(\text{attr}_k(\mathcal{O}_{i,j}^{act})|\mathcal{P}).
    \end{equation}

    \stitle{Adaptive Calculation Strategies.}
        Given the complex scenarios in operation recommendations, relying on a static formula to calculate the confidence score is unreliable. Therefore, we introduce several adaptive strategies to refine this process:
        1) The confidence threshold is dynamically adjusted according to task complexity. It is lowered by 10\% for predictions with larger candidate pools or for numerical attributes (e.g., temperature) from a continuous range, accounting for the inherent difficulty of these tasks.
        2) To enhance efficiency, we employ a cascading pruning mechanism. An entire recommendation is discarded if the confidence for a high-level attribute, such as the device, falls below its threshold. This hierarchical check prevents wasting computational resources on low-confidence candidates. 
        3) To create a more robust and reliable metric, the final confidence score is a fusion of the average confidence across all actions and the confidence of the first-generated token. This prevents scenarios where a high average score might mask low confidence in the foundational first token, which is often critical for the entire recommendation's validity.
